\pdfoutput=1
\def\compileforpublish{1}
\def\isaccepted{1}

\documentclass[letterpaper, 10 pt, conference]{ieeeconf}  

\IEEEoverridecommandlockouts                              

\makeatletter
\newcounter{IEEE@bibentries}
\renewcommand\IEEEtriggeratref[1]{%
	\renewbibmacro{finentry}{%
		\stepcounter{IEEE@bibentries}%
		\ifthenelse{\equal{\value{IEEE@bibentries}}{#1}}
		{\finentry\@IEEEtriggercmd}
		{\finentry}%
	}%
}
\makeatother

\ifx \isaccepted \undefined
\newcommand\copyrighttext{%
	\footnotesize \centering This work has been submitted to the IEEE for possible publication.\\ Copyright may be transferred without notice, after which this version may no longer be accessible.}
\else
\newcommand\copyrighttext{%
	\footnotesize \parbox[t]{.11\textwidth}{\copyright{} 2017~IEEE.} \parbox[t]{.87\textwidth}{Personal use of this material is permitted. Permission from IEEE must be obtained for all other uses, in any current or future media, including reprinting/republishing this material for advertising or promotional purposes, creating new collective works, for resale or redistribution to servers or lists, or reuse of any copyrighted component of this work in other works.}}
\fi
\newcommand\copyrightnotice{%
	\ifx \compileforpublish \undefined
	\else
	\begin{tikzpicture}[remember picture,overlay]
	\node[anchor=south,yshift=10.5pt] at (current page.south) {\parbox{\dimexpr\textwidth-\fboxsep-\fboxrule\relax}{\copyrighttext}};
	\end{tikzpicture}%
	\fi
}

\usepackage{tikz,amsmath, amssymb,bm,color,graphicx,siunitx}
\usepackage{pdflscape}
\usepackage{url}
\usepackage{array,booktabs,colortbl}
\usepackage{siunitx} 
\usepackage{tikz}
\usepackage{longtable}
\usepackage{multicol}
\usepackage{multirow}
\usepackage{threeparttable}
\usepackage{capt-of}
\usepackage{pgfplotstable}  
\usetikzlibrary{arrows, shapes, positioning, shapes.geometric, decorations.markings, calc, fit}

\pgfplotsset{compat=1.13}

\definecolor{lightgrey}{gray}{0.9}

\newcommand{\eg}{e.\,g.}
\newcommand{\ia}{i.\,a.}
\newcommand{\afalogic}{{AFA Logic}}
\newcommand{\hara}{{\textsc{HARA}}}
\newcommand{\afas}{aFAS}

\newcommand{\followmode}{\emph{Follow Mode}}
\newcommand{\safehalt}{\emph{Safe Halt}}
\newcommand{\coupledmode}{\emph{Coupled Mode}}
\newcommand{\manualmode}{\emph{Manual Mode}}

\title{\LARGE \bf
	Hazard Analysis and Risk Assessment for an Automated 
	Unmanned Protective Vehicle*}

\author{Torben Stolte$^{1}$, Gerrit Bagschik$^{1}$, Andreas Reschka$^{1}$, and Markus Maurer$^{1}$
\thanks{*The project \afas\ is partially funded by the German Federal Ministry of Economics and Technology (BMWi).
	The project consortium consists of MAN (consortium leader), ZF TRW, WABCO, Bosch Automotive Steering,  Technische Universit\"at Braunschweig, Hochschule Karlsruhe, Hessen Mobil~- Road and Traffic Management, and BASt - Federal Highway Research Institute. 
	We would like to thank our partners and colleagues for their support of our work as well as the reviewers for the high quality review.
}
\thanks{$^{1}$Torben Stolte, Gerrit Bagschik, Andreas Reschka, and Markus Maurer are
	with the Institute of Control Engineering at Technische Universit\"at Braunschweig, 38106 Braunschweig, Germany.
    {\tt\small \{stolte, bagschik,reschka,maurer\}@ifr.ing.tu-bs.de}}%
}

\begin{document}

\maketitle
\thispagestyle{empty}
\pagestyle{empty}

\begin{abstract}
For future application of automated vehicles in public traffic, ensuring functional safety is essential. 
In this context, a hazard analysis and risk assessment is an important input for designing functionally vehicle automation systems. 
In this contribution, we present a detailed hazard analysis and risk assessment (HARA) according to the ISO~26262 standard for a specific Level~4 application, namely an unmanned protective vehicle operated without human supervision for motorway hard shoulder roadworks. 
\end{abstract}

\setcounter{footnote}{1}

\section{Introduction}
The automation of the driving task is probably the most challenging field of research in the automotive context. 
Level~4 and Level~5 systems --~according to the definition of SAE~\cite{sae_2016}~-- combine the unlimited set of operational scenarios encountered in public traffic with the absence of human supervision.
This implies highest demands regarding functional safety throughout the development of these systems.
Thus, the applicability of the ISO 26262 standard~\cite{iso_2011} --~the most recent standard for designing safety-relevant electronic systems in the automotive context~-- must be examined. 

Following the ISO~26262 standard, a hazard analysis and risk assessment (\hara) is required in order to determine the criticality of the system under consideration. 
The results of the \hara\ strongly influence the efforts to be undertaken in the subsequent development steps for ensuring functional safety. 
Normally, the results of {\hara}s are not published and thus cannot be discussed in the scientific community due to reasons of non-disclosure. 
This also applies to the field of vehicle automation. 

However, exceptionally high demands regarding system implementation and its safety result from the missing human supervision. 
Hence, in-depth discussions about functional safety are crucial before deploying automated vehicles in public traffic. 
In this contribution, we present the complete results of a \hara\ conducted for a specific Level~4 application. 
The paper structures as follows: We introduce the project \afas\ and 
the functionality to be implemented in the project in Section~\ref{sec:systemdescription}. 
In Section~\ref{sec:hara}, we define relevant terms, describe the \hara\ approach, and highlight important results. 
Finally, Section \ref{sec:evaluation} contains 
the implications on designing vehicle automation systems. 
Complete results of the conducted \hara\ can be taken from the Appendix.
\copyrightnotice
\vspace{-.1cm}
\section{System Description \& Project Context}
\label{sec:systemdescription}

The project \afas
	\footnote{German abbreviation for ``Automated Unmanned Protective Vehicle for Highway Hard Shoulder Road Works''} 
aims at developing an unmanned operation of a protective vehicle (AFA
	\footnote{German abbreviation for ``Automated Unmanned Protective Vehicle''}
) on the hard shoulder of highways in Germany, cf. \cite{stolte_2015}. 
The vehicle is operated without supervision on hard shoulders only and with low speed of up to \SI{12}{\km\per\hour}.
The automated operation consists of three operating modes complemented by the \manualmode\ which comprises the normal operation of the AFA with a human driver. 
\safehalt\ serves as initial operating mode as well as for switching between \followmode\ and \coupledmode. 
Furthermore, \safehalt\ is activated if the system leaves functional system boundaries. 
In \followmode, the AFA follows the leading vehicle, which conducts the actual work such as cleaning the hard shoulder, in a defined distance of about \SI{90}{\meter}. 
To follow the leading vehicle and to stay on the hard shoulder, the AFA perceives the leading vehicle as well as lane markings of the hard shoulder by environment sensors.
In \coupledmode, the AFA follows the leading vehicle in close distance of about \SI{10}{\meter} in order to pass acceleration and deceleration lanes. 
This is primarily realized through motion data of the leading vehicle.
For transmitting system states and commands (\eg\ changes of operating modes), the vehicles communicate via radio. 

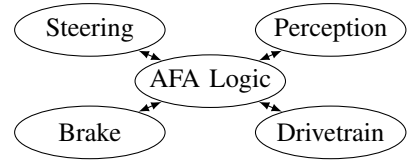
\begin{figure}
	\centering
	\tikzstyle{bubble}=[draw, ellipse, minimum width=2cm, minimum height=.7cm]

\begin{tikzpicture}
\node[bubble] (afalogic) {};
\node at (afalogic) {AFA Logic};
\node[bubble, above left=.25cm of afalogic] (steering) {};
\node at (steering) {Steering};
\node[bubble, below left=.25cm of afalogic] (brake) {};
\node at (brake) {Brake};
\node[bubble, below right=.25cm of afalogic] (drive) {};
\node at (drive) {Drivetrain};
\node[bubble, above right=.25cm of afalogic, align = center] (perception) {};
\node at (perception) {Perception};

\draw[latex-latex] (afalogic) to (brake);
\draw[latex-latex] (afalogic) to (steering);
\draw[latex-latex] (afalogic) to (drive);
\draw[latex-latex] (afalogic) to (perception);
\end{tikzpicture}
	\caption{Dependence of AFA Logic and connected elements}
	\label{fig:items}
	\vspace{-.65cm}
\end{figure}

For the \hara\ presented in the following (cf. Section~\ref{sec:hara}), 
we concentrated on the parts that are specific for the automated operation in order to reduce the complexity that arises when considering the entire vehicle. 
The considered functionality is summarized in terms of the item
	\footnote{Defined as ``system or array of systems to implement a function at the vehicle level, to which ISO 26262 is applied''~\cite[1.69]{iso_2011}} 
called \afalogic. 
However, for unmanned operation additional elements are required, namely drivetrain, brakes, steering and environment perception. 
These elements are connected with the \afalogic\ as depicted in Fig.~\ref{fig:items}. 
Hence, safety requirements can be inherited between connected elements.

\section{Hazard Analysis and Risk Assessment}
\label{sec:hara}

\subsection{Terminology}
\label{subsec:terms}
A major contribution of the ISO~26262 standard is the definition of more than 100 terms related to functional safety of automotive electric/electronic systems. 
Yet, some terms must be further clarified for automated driving. 
In the context of the \hara, the terms \emph{hazard}, \emph{hazardous event}, \emph{operational situation}, and \emph{malfunctioning behavior} are the most common terms encountered. 
The term \emph{hazard} \cite[1.57]{iso_2011} is defined as ``potential source of harm'', which is consistent to other definitions in safety engineering, \eg\ \cite{iec_2010,iso_2014}.
The definition used in the ISO~26262 standard specifies that a hazard is caused by \emph{malfunctioning behavior}. 
\emph{Malfunctioning behavior} itself is either caused by failures or unintended behavior of the system~\cite[1.73]{iso_2011}. 
Hence, the definitions of \emph{hazard} and \emph{malfunctioning behavior} are applicable for automated driving.

Furthermore, combining \emph{operational situation} and \emph{hazard} yields a \emph{hazardous event} \cite[1.59]{iso_2011}. 
In contrast to \emph{hazard} and \emph{malfunctioning behavior}, the ISO~26262 standard's definitions of the terms \emph{operational situation} and \emph{hazardous event} are vague with respect to automated driving. 
A similar vagueness can be found in \cite{iec_2010} and \cite{iso_2014}.
An \emph{operational situation} is defined as ``scenario that can occur during a vehicle's life'' \cite[1.83]{iso_2011}, equaling the terms \emph{situation} and \emph{scenario}. 
However, both terms --~together with the term \emph{scene}~-- are widely used in the context of automated driving and must be distinguished from each other according to Ulbrich et al.~\cite{ulbrich_2015}, who present a comprehensive literature review regarding these terms. 
Ulbrich et al. define and substantiate a \emph{scene} as an all-encompassing snapshot of an environment together with the self-representation of all actors and observers contained (objective \emph{scene}). 
In the real world, a scene is always subjective for each observer. 
A \emph{situation} is derived from the subjective \emph{scene} perceived by a traffic participant.
It contains all necessary premises to derive suitable driving decisions. 
A \emph{scenario} is the temporal concatenation of related \emph{scenes}. 
Hence, we utilize the term \emph{operational scenario} in preference to \emph{operational situation} in the following since an objective exterior view is what is required for conducting a \hara.

The vagueness of the term \emph{hazardous event} results from the linguistic ambiguity of the term \emph{event}.
This ambiguity is not resolved in the ISO~26262 standard. 	
\emph{Event} either addresses a period of time or -- in a physical/technical sense -- a point of time \cite{oxforddictionaries_}. 
In engineering, one would consider the latter as intended meaning, yet the temporal interpretation is meant by the ISO~26262 standard in our understanding. 
What is actually required for obtaining a classification of safety criticality, is an \emph{operational scenario} combined with a \emph{hazard}. 
Thus, we utilize the term \emph{hazardous scenario} in preference to \emph{hazardous event} in the following.

\vspace{-.15cm}
\subsection{Approach}
\label{subsec:approach}

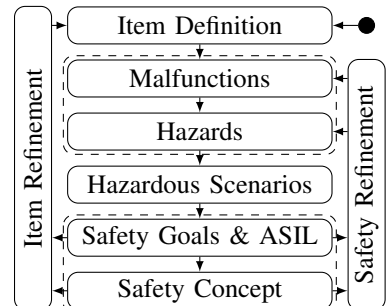
\begin{figure}[b]
	\vspace{-.5cm}
	\centering
	\tikzstyle{box}=[draw, rounded corners, minimum height=14, minimum width=100]

\begin{tikzpicture}[x=1mm,y=1mm]

\node[box] (itemdefinition){};
\node [] at (itemdefinition) {Item Definition};
\node[box, below=2 of itemdefinition](malfunction){};
\node [] at (malfunction) {Malfunctions};
\node[box, below=2 of malfunction](hazard){};
\node [] at (hazard) {Hazards};
\node[box, below=2 of hazard](hazScenario){};
\node [] at (hazScenario) {Hazardous Scenarios};
\node[box, below=2 of hazScenario](safetygoals){};
\node [] at (safetygoals) {Safety Goals \& ASIL};
\node[box, below=2 of safetygoals](safetyconcept){};
\node [] at (safetyconcept) {Safety Concept};
\path   let	\p1=($(itemdefinition.north)-(safetyconcept.south)$),
		 	\n1={veclen(\x1,\y1)} 
		in	(safetyconcept.south) node[box, rotate = 90, left =2 of $0.5*(hazard.west)+0.5*(hazScenario.west)$, anchor = south, minimum width=\n1](refinement){};
\node [rotate=90] at (refinement){Item Refinement};
\path   let	\p1=($(malfunction.north)-(safetyconcept.south)$),
		 	\n1={veclen(\x1,\y1)} 
		in	(safetyconcept.south) node[box, rotate = 90, right =2 of hazScenario.east), anchor = north, align=center, minimum width=\n1](SGrefinement){};
\node [rotate=90] at (SGrefinement){Safety Refinement};

\node [draw,circle,fill,minimum size=6,inner sep =0 ] at (itemdefinition-|SGrefinement) (init){};

\draw[-latex] (itemdefinition)	to (malfunction);
\draw[-latex] (malfunction) 		to (hazard);
\draw[-latex] (hazard)			to (hazScenario);
\draw[-latex] (hazScenario)		to (safetygoals);
\draw[-latex] (safetygoals)		to (safetyconcept);
\draw[-latex] (safetygoals)		to (refinement.south|-safetygoals);
\draw[-latex] (safetyconcept)	to (refinement.south|-safetyconcept);
\draw[latex-] (itemdefinition)	to (refinement.south|-itemdefinition);
\draw[-latex] (safetygoals)		to (SGrefinement.north|-safetygoals);
\draw[-latex] (safetyconcept)	to (SGrefinement.north|-safetyconcept);
\draw[latex-] (malfunction)		to (SGrefinement.north|-	malfunction);
\draw[latex-] (hazard)			to (SGrefinement.north|-	hazard);
\draw[-latex] (init)				to (itemdefinition);

\draw[rounded corners,dashed]  ($(malfunction.north west)+(-.5,.5)$) rectangle ($(hazard.south east)-(-.5,.5)$);
\draw[rounded corners,dashed]  ($(safetygoals.north west)+(-.5,.5)$) rectangle ($(safetyconcept.south east)-(-.5,.5)$);
\end{tikzpicture}
	\vspace{-.3cm}
	\caption{Process of \hara\ generation and refinement}
	\label{fig:haraprocess}
\end{figure}

For conducting a \hara, a linear reference process is illustrated in the ISO~26262 standard \cite[Part 3]{iso_2011}, which rather addresses the interdependencies of single steps than necessary iterations to reach completeness. 
Warg et~al.~\cite{warg_2016} describe an iterative process for developing \hara s in the context of automated driving
. The process we applied in regard to the \afalogic\ is similar to the approach proposed by Warg et al. and is depicted in Fig.~\ref{fig:haraprocess}.
Yet, our approach differs from the approach of Warg et al. in certain aspects and extends it as well: 
While Warg et al. take a \emph{preliminary feature description} as initial input resulting in an item definition during the process, our process input is a well advanced item definition. 

Furthermore, we introduce two loops instead of one for refining the work products. 
Effects of both loops on the \afalogic\ are described in the following subsection.
The \emph{item refinement} --~comparable to the function refinement of Warg et al.~\cite{warg_2016}~-- describes extending or (in most cases) narrowing the functional range of the item under consideration.
By this means, the merely functional consideration of the item according to the ISO~26262 standard is supplemented by considering technical feasibility, \eg\ due to limited project resources or not yet available technology. 
In contrast, the \emph{safety refinement} does not affect the functional range. 
Rather, it aims at refining the determined hazardous scenarios in order to enable technically realizable safety concepts through reaching more precise and definite safety goals. 
The {safety refinement} is comparable to the procedure to reach completeness of \hara s of Johansson~\cite{johansson_2015}.

Apart from refinement, each iteration loop consists of six steps. 
In the first step, functionalities are extracted from the item definition. Subsequently, potential malfunctioning behavior and related hazards are derived in the second and third step, respectively.  
The combination of hazards and operational scenarios derived from the item definition then yields the hazardous scenarios in the fourth step.
Determining \emph{Automotive Safety Integrity Levels} (ASILs) and safety goals are the fifth and the sixth step in Fig.~\ref{fig:haraprocess}, which are strongly linked.

\vspace{-.2cm}
\subsection{Results}
\label{subsec:results}
We developed the \hara\ together with experts from the industrial members of the \afas\ consortium$^1$, in iterative group meetings. 
As mentioned, the complete results can be found in the Appendix. 
In the following, we highlight selected results that affect functional range, environment perception, human machine interface (HMI), and user interaction, as well as central control logic. 
Table \ref{tab:SafetyGoals} presents the identified safety goals. 
Its order --~as the \hara's numbering in the Appendix~-- illustrates the several iterations necessary to obtain a result commonly accepted among the contributors. 
For instance, safety goal SG16 was added in a later iteration, although strongly connected to safety goal SG04. 

\newlength{\SGIDwidth}
\setlength{\SGIDwidth}{.625cm}
\newlength{\SGwidth}
\setlength{\SGwidth}{\columnwidth}
\addtolength{\SGwidth}{-\SGIDwidth}
\addtolength{\SGwidth}{-4.0\tabcolsep}

\setlength{\abovetopsep}{0pt}
\setlength{\belowrulesep}{0pt}
\setlength{\aboverulesep}{0pt}
\setlength{\belowbottomsep}{0pt}
\setlength{\defaultaddspace}{.75mm}

\begin{table}
	\caption{Protective Vehicle's Safety Goals for Unmanned Operation}
	\label{tab:SafetyGoals}
	\vspace{-.5cm}
	\begin{center}
	\begin{tabular}{>{\centering\arraybackslash}m{\SGIDwidth}>{\centering\arraybackslash}m{\SGwidth}}
	\toprule
	ID	& Safety Goal\\
	\midrule
	\addlinespace 
	\multicolumn{2}{c}{{All operating modes}}\\
	\midrule
	\rowcolor{lightgrey}SG01&Unintended and not permitted operating mode change must be prevented.\\
	SG02&Intended and permitted operating mode change must be ensured.\\
	\rowcolor{lightgrey}SG07&Display of actual operating mode in HMI must be ensured.\\
	\midrule
	\addlinespace 
	\multicolumn{2}{c}{{Manual Mode}}\\
	\midrule
	\rowcolor{lightgrey}SG04&Unintended anti-lock brake actuation must be prevented.\\
	SG05&Unintended acceleration must be prevented.\\
	\rowcolor{lightgrey}SG16&Anti-lock functionality must be ensured.\\
	SG17&Unintended steering actuation must be prevented\\
	\midrule
	\addlinespace 
	\multicolumn{2}{c}{{Follow Mode, Coupled Mode, Safe Halt}}\\
	\midrule
	\rowcolor{lightgrey}SG03&Steering actuation beyond specification must be prevented.\\
	SG06&Detection of driver intervention must be ensured.\\
	\rowcolor{lightgrey}SG08&Unintended slow acceleration must be prevented.\\
	SG09&Deceleration to standstill must be ensured.\\
	\rowcolor{lightgrey}SG10&Leaving tolerance ranges must trigger operating mode change to Safe Halt.\\
	SG11&Maximum velocity must not be exceeded.\\
	\rowcolor{lightgrey}SG12&Overrunning hard shoulder markings must be prevented.\\
	SG13&Detection of and reaction to (deceleration to standstill) relevant obstacles (humans, vehicles, etc.) must be ensured. \\
	\rowcolor{lightgrey}SG14&Identification of leading vehicle must be ensured.\\
	SG15&Detection of missing leading vehicle and operating mode change to safe halt must be ensured.\\
	\bottomrule
	\end{tabular}
	\end{center}
\vspace{-1cm}	
\end{table}

Initially, the functional range was supposed to include automated unmanned operation on the motorway's right lane as well, in order to be capable of driving around obstacles on the hard shoulder. 
During \emph{item refinement}, however, the functional range was reduced, as this feature was technically too challenging due to limited project resources. 
Accordingly, the unmanned operation was restricted to hard shoulders as well as acceleration and deceleration lanes, both with a limited velocity of \SI{10}{\km\per\hour} (plus \SI{2}{\km\per\hour} tolerance). 

An example for the above mentioned \emph{safety refinement} loop are safety goals SG03 and SG12. 
Safety goal SG12 was established in one of the first iterations effecting high ASIL ratings on all involved components, namely environment perception, \afalogic, and actuators. 
In subsequent iterations, we differentiated between unintended steering actuation beyond the specification of the item definition (up to full steering actuation) and unintended steering actuation within the specification. 
This results in different hazardous scenarios which were rated separately. 
Consequently, safety goal SG03 was introduced, which targets at limiting the maximum steering angle and thereby reduces the effects of malfunctioning behavior of other system elements. 
Due to the limited steering angle, the AFA will intrude the right lane of the motorway with less lateral velocity. 
Thus, the controllability rating can be reduced as other traffic participants can react more appropriately.
By this means, the limitation of the steering angle in automated operation gains the former high ASIL rating (ASIL~D) of safety goal SG12 while the rating of safety goal SG12 is reduced (ASIL~B).

The previous reduction of the ASIL rating of safety goal SG12 also affects the functional block of the \afalogic, which must be implemented with ASIL~B as well. 
In discussions prior to the project start, a group of experts from the consortium underestimated the efforts to be undertaken for implementing the \afalogic\ 
as well as of the human machine interaction. 
If the operating mode is wrongly displayed, the AFA could intrude the right lane of the motorway and cause severe accidents, cf. \hara\ IDs 37 and~37a in the Appendix. 
Consequently, the correct display of the actual operating mode must be ensured (safety goal SG07, ASIL~A). 
Both aspects illustrate the high demands on all system parts which originate from the missing human supervision. 

The \hara's results concerning the environment perception are of particular interest, since automated vehicles are operated in an open environment where they encounter an infinite set of operational scenarios.
For the AFA, safety goals SG12 and SG13 address environment perception. 
While safety goal SG12 obtained an ASIL~B rating, the detection and reaction to obstacles on the path are rated with QM since persons involved in the scenarios can generally control the scenarios due to the low velocity of the AFA. 

As already depicted in Fig.~\ref{fig:items}, the \afalogic\ is connected to further items. 
Although a \hara\ is a top-down procedure, at some points technical aspects must be considered. 
In the planned system implementation of the unmanned operation, the \afalogic\ has access to steering and brakes. 
In particular, the technical implementation of the brake system creates potential for malfunctioning behavior. 
Therefore, the manual operation must be considered in the \hara\ as well. 
As the malfunctioning behavior can create critical outcome in several scenarios, the related safety goals obtain highest ASIL ratings. 
This means that elements connected to the \afalogic\ inherit according safety requirements.

\vspace{-.15cm}

\section{Discussion and Related Work}
\label{sec:evaluation}
Although the functional range considered in the \afas\ project is small compared to functional ranges of future automated vehicles, several implications can be derived from our experiences made. 
The presented \hara\ is primarily based on the experience and knowledge of the involved contributors from industry and academia with a range of experience from one to more than ten years. 
Despite the small functional range, the contributors agree on that it was challenging to take all relevant aspects into account in order to reach consistency between item definition and \hara. 
This reflects in the several iterations necessary to reach a common result. 
Using only expert knowledge might lead to missed scenarios and thus to building unsafe systems. 
Consequently, we expect that HARAs for systems featuring more comprehensive functional ranges must be supported by methods and tools. 
The approach for refining item and safety aspects described in subsection \ref{subsec:approach} appears suitable in general. 
However, more distinguished methods must be developed for single steps in order to gain appropriate results. 

As input to the \hara\ process, the \afalogic's item definition is written in natural language, supported by some tables and figures. 
All functionalities considered in the HARA were extracted manually. 
This was a process taking several iterations since functionalities had not been considered or had initially been defined contradictory.  
For items with a wider functional range, item definitions with a more extensive utilization of semi-formal or even formal notations are necessary for ensuring proper identification of all relevant functions and related malfunctioning behavior. 
Moreover, this eases traceability between item definition and \hara.

For targeting completeness of hazardous scenarios, different approaches for identifying hazards and operational scenarios can be found in literature. 
Comparable to the approach in the \afas\ project, Johansson \cite{johansson_2015} suggests experts to challenge each single hazardous scenario.
If they do not find additional scenarios that lead to new safety goals, the list is likely to be complete states Johansson. 
However, correct ASIL ratings are required besides completeness of safety goals. 
Thus, the \afas\ consortium also considered ASIL ratings of hazardous scenarios with the same safety goals.
Warg et al.~\cite{warg_2016} propose an identification of both hazards as well as operational scenarios based on tree structures. 
Out of the \afas\ consortium, Bagschik et~al.~\cite{bagschik_2016a} propose an approach for deriving all relevant hazardous scenarios systematically by combining operating modes, functions (derived by skill graphs), malfunctions (derived by a HAZOP analysis), and scene discretization. 
However, suitability of these approaches still needs to be proven for systems of future automated vehicles in terms of considering all relevant scenarios. 
The first two approaches need to prove their suitability for automated vehicles with a wider functional range. 
In contrast, the approach of Bagschik et al. creates automatically an extensive list of scenarios. However, each scenario must be assessed manually regarding safety criticality. 

Once hazardous scenarios are identified, the next challenge is determining the ASIL classification. 
As already mentioned, the classification for the unmanned operation of the AFA is based on expert knowledge. 
A few aspects of the exposure -- such as the rate of emergency stopping vehicles -- are justified by investigations of the \afas\ consortium.  
Severity and controllability are purely based on experts' contribution. 
Furthermore, standards such as the SAE~J2980 standard~\cite{sae_2015} are of limited contribution for the project \afas\, since they do not consider operations on the hard shoulder and focus on vehicle motion control systems.
In general, controllability of hazardous scenarios is very low for Level~4 or Level~5 applications with passengers. 
The controllability  of hazardous scenarios without passengers --~as in the project \afas~-- is determined by surrounding traffic participants.
For future application of automated vehicles, methods for objectification of the parameters must be discussed. 
At least, evolving standards such as the SAE~J2980 standard~\cite{sae_2015} towards automated driving can support a common understanding. 

So far, we conclude that methods for a systematic consideration of each \hara\ step can be found in literature. 
Consequently, one can argue that a holistic systematic \hara\ process is beneficial, as \ia\ presented by Kemmann and Trapp~\cite{kemmann_2011} as well as by Beckers et al.~\cite{beckers_2013}. 
Kemmann and Trapp \cite{kemmann_2011} introduce 
\emph{A Structured Approach for Hazard Analysis and Risk Assessments} (SAHARA), which systematically considers each \hara\ step. 
The authors consequently use model based approaches for item definition, hazard identification, as well as for classification of controllability, severity, and exposure. 
Beckers et al.~\cite{beckers_2013} emphasize utilization of UML based notation. 
This ensures the traceability throughout the \hara\ process and enables potential for formal verification. 
Still, single \hara\ steps in the approach of Becker et al. strongly depend on expert knowledge. 
For both approaches, proof of applicability to automated vehicles must be furnished.

\vspace{-.25cm}

\section{Conclusion}

The example of the unmanned protective vehicle 
reveals challenges during a \hara\ for automated vehicles operated without human supervision. 
It was demonstrated that conventional \hara\ approaches are of limited suitability, especially for future applications with a wider functional range. 
Consequently, already existing systematic approaches must be evolved towards automated driving functionalities without human supervision. 
For this, an in-depth consideration of each single \hara\ step is required.
Furthermore, for merging the two worlds of automated driving and functional safety, clarification of used terminology is crucial to reach a common understanding.

\vspace{-.15cm}

\bibliographystyle{IEEEtran}
\bibliography{literature}

	\onecolumn
\begin{landscape}
	\appendix
	\newlength{\IDwidth	}
\setlength{\IDwidth}{.24cm}
\newlength{\modeWidth}
\setlength{\modeWidth}{.7cm}
\newlength{\functionWidth}
\setlength{\functionWidth}{.9cm}
\newlength{\malfunctionWidth}
\setlength{\malfunctionWidth}{1.75cm}
\newlength{\hazardWidth}
\setlength{\hazardWidth}{.6cm}
\newlength{\SECWidth}
\setlength{\SECWidth}{.19cm}
\newlength{\SRationaleWidth}
\setlength{\SRationaleWidth}{2.5cm}
\newlength{\ERationaleWidth}
\setlength{\ERationaleWidth}{2.5cm}
\newlength{\CRationaleWidth}
\setlength{\CRationaleWidth}{3.5cm}
\newlength{\ASILWidth}
\setlength{\ASILWidth}{.27cm}
\newlength{\SGWidth}
\setlength{\SGWidth}{.39cm}
\newlength{\hazardousEventWidth}
\setlength{\hazardousEventWidth}{9.5in}
\addtolength{\hazardousEventWidth}{2\tabcolsep}
\addtolength{\hazardousEventWidth}{-\IDwidth}
\addtolength{\hazardousEventWidth}{-\modeWidth}
\addtolength{\hazardousEventWidth}{-\functionWidth}
\addtolength{\hazardousEventWidth}{-\malfunctionWidth}
\addtolength{\hazardousEventWidth}{-\hazardWidth}
\addtolength{\hazardousEventWidth}{-3.0\SECWidth}
\addtolength{\hazardousEventWidth}{-\SRationaleWidth}
\addtolength{\hazardousEventWidth}{-\ERationaleWidth}
\addtolength{\hazardousEventWidth}{-\CRationaleWidth}
\addtolength{\hazardousEventWidth}{-\ASILWidth}
\addtolength{\hazardousEventWidth}{-\SGWidth}
\addtolength{\hazardousEventWidth}{-26.0\tabcolsep}

\newcommand{\bottomtext}{
	\multicolumn{10}{l}{{ID: Identifier for hazardous scenario as in original document, 
			S: Severity* (S0--S3), 
			E: Exposure* (E0--E4), 
			C: Controllability* (C0--C3), 
			A: ASIL Rating (QM, ASIL A--D), 
			SG: ID of Safety Goal (cf. Table~\ref{tab:SafetyGoals}); 
			*For reference cf. \cite[Part 3]{iso_2011} and \cite{sae_2015}}}
}

\setlength{\belowrulesep}{0pt}
\setlength{\aboverulesep}{0pt}
Table \ref{tab:HARA} displays the \hara\ developed in the project \afas. Omitted and alphanumeric IDs reflect the iterative process of \hara\ development during item and safety refinement, cf. subsection \ref{subsec:approach}.  
Several IDs were discarded while the ID numbering was not adjusted, in order to preserve traceability between different \hara\ versions.
\vspace{-.35cm}

\captionof{table}{Results of Hazard Analysis and Risk Assessment of the Automated Operation of the Unmanned Protective Vehicle}
	\label{tab:HARA}
\vspace{-.5cm}
\pgfkeysifdefined{/pgfplots/table/output empty row/.@cmd}{
	\pgfplotstableset{
		empty header/.style={
			every head row/.style={output empty row},
		}
	}
}{
	\pgfplotstableset{
		empty header/.style={
			typeset cell/.append code={%
				\ifnum\pgfplotstablerow=-1 %
				\pgfkeyssetvalue{/pgfplots/table/@cell content}{}%
				\fi
			}
		}
	}
}
\addtocounter{table}{-1}
\tiny
\begin{filecontents*}{tables/hara.csv}
Original ID;Consecutive ID;Operating mode;Function;Malfunction;Preliminary Assessment;Hazard;Hazardous Event;Severity;Severity Rationale;Exposure;Exposure Rationale;Controllability;Controllability Rationale;ASIL Classification;Safety Goal ID;Safety Goal
1;1;Manual Mode;Operating mode change;Unintended or not permitted transition to other operating mode;Hazard;Collision with traffic participants;Drive on road in convoy (up to \SI[mode=text]{90}{\km\per\hour}, depending on speed limiter). Operating mode change to \safehalt\ leads to unpredictable deceleration  (\SI[mode=text]{4}{\metre/\square\second} in \safehalt\ according to Item Definition) to standstill. The same applies when changing to \coupledmode\ or \followmode, as these operating modes then will be operated beyond accepted parameters which causes an operating mode change to \safehalt. Deceleration leads to rear-end collision of succeeding vehicle.;S2;Assuming succeeding traffic participants use seat belts and brakes intuitively, the collision will happen with medium velocity which leads to severe injuries.;E4;Drive to and from location of road works via roads and motorways occurs for each road work.;C2;Traffic participants not complying with traffic rules is commonly observable on German motorways (exceeded velocities, tailgating, etc.). Reaction time \SI[mode=text]{1.5}{\second}, distance~\SI[mode=text]{35}{\metre}, deceleration \SI[mode=text]{-6}{\metre/\square\second}. ;B;SG01;Unintended and not permitted operating mode change must be prevented.
2;2;Manual Mode;Steer;Unintended steering;Hazard;Collision with traffic participants;Drive on road or motorway. Unpredictable swerving from lane leads to collision with other traffic participants.;S3;Road: Head-on collision with oncoming traffic, tree etc. motorway: Collision with moving traffic. Both scenarios can lead to severe or fatal injuries.;E4;Drive to and from location of road works via roads and motorways occurs for each road work.;C3;Experience of Bosch Automotive Steering: Full steering angle due to failures is not controllable.;D;SG17;Unintended steering actuation must be prevented in \manualmode\
3;3;Manual Mode;Brake;Unintended braking with anti-lock functionality;Hazard;Collision with traffic participants;Drive on road in convoy. Unpredictable maximum deceleration with anti-lock functionality leads to rear-end collision of succeeding vehicle. Malfunction possible due to the planned technical implementation.;S2;Assuming succeeding traffic participants use seat belts and brakes intuitively, the collision will happen with medium velocity which leads to severe injuries.;E4;Drive to and from location of road works via roads and motorways occurs for each road work.;C3;Traffic participants not complying with traffic rules is commonly observable on German motorways (exceeded velocities, tailgating, etc.). Reaction time \SI[mode=text]{1.5}{\second}, distance~\SI[mode=text]{35}{m}, deceleration \SI[mode=text]{-6}{\metre/\square\second};C;SG04;Unintended anti-lock brake actuation must be prevented.
4;4;Manual Mode;Brake;Unintended braking without anti-lock functionality;Hazard;Collision with traffic participants;Drive on road or motorway. Unpredictable maximum deceleration without anti-lock functionality leads to locking tires. Lateral guidance is not possible, the AFA becomes uncontrollable. The AFA leaves its lane and collides with stationary objects or other vehicles.  Malfunction possible due to the planned technical implementation.;S3;Collision with uncontrollable vehicle at high velocities leads to severe or fatal injuries. ;E4;Drive to and from location of road works via roads and motorways occurs for each road work.;C3;According to  ISO26262-3, Table B4: Failure of brakes $\rightarrow$ brakes unintendedly stopping the vehicle;D;SG16;Anti-lock functionality must be ensured.
5;5;Manual Mode;Drive;Unintended acceleration;Hazard;Collision with traffic participants;Drive on road in convoy. Unintended acceleration leads to rear-end collision with preceding vehicle. ;S3;Traffic participant skids, resulting crash leads to severe and life-threatening injuries.;E4;Drive to and from location of road works via roads and motorways occurs for each road work.;C0;Controllable in general. Due to inertia enough time for driver of AFA to react, driver brakes intuitively.;QM;SG05;Unintended acceleration must be prevented.
#06;6;Manual Mode;HMI;HMI displays automated operating mode;No Hazard;; ;S0;Only road workers with special training are deployed on AFA.;E0;Only road workers with special training are deployed on AFA.;C0;Only road workers with special training are deployed on AFA.;QM; ; 
7;7;Manual Mode;HMI;HMI displays wrong operating mode;No Hazard;;\nohazard{Only road workers with special training are deployed on AFA. In \manualmode, the AFA is driven as usual. A wrong display of operating modes leads to no more than short confusion.};S0;---;E0;---;C0;---;QM;---;
8;8;Safe Halt;Detection of driver intervention;Driver intervention is not detected;No Hazard; ;Test operation on hard shoulder, driver intervention is not detected. ;S0;Driver intervention not detected in \safehalt\ does not lead to a hazardous event.;E0;Driver intervention not detected in \safehalt\ does not lead to a hazardous event.;C0;Driver intervention not detected in \safehalt\ does not lead to a hazardous event.;QM;SG06;Detection of driver intervention must be ensured.
9;9;Safe Halt;HMI;HMI displays wrong operating mode;No Hazard; ;\nohazard{AFA and leading vehicle in standstill. Operating mode change can only be triggered by operator in leading vehicle.};S0;---;E0;---;C0;---;QM;---; 
10;10;Safe Halt;Operating mode change;Unintended or not permitted transition to \manualmode;Hazard;Collision with traffic participants;Truck convoy on right lane.  AFA on hard shoulder starts to roll (slope, automatic gearbox). This leads to unpredictable behavior including intrusion into right lane. Truck avoids AFA, following truck touches AFA as the AFA is masked by first truck.;S2;Collision with high differential velocity, vehicles slightly touch.;E4;Scenario (sloped road, traffic on right lane) usually met at each deployment.;C2;AFA drifts slowly (\eg\ \SI[mode=text]{0.4}{\metre/\second} lateral) into right driving lane. Despite of masking by truck in front, following traffic is normally able to recognize this and react appropriately (braking, avoiding).;B;SG01;Unintended and not permitted operating mode change must be prevented.
10a;11;Safe Halt;Operating mode change;Unintended or not permitted transition to \manualmode;Hazard;Collision with traffic participants;Moving car traffic on right lane. AFA on hard shoulder starts to roll (slope, automatic gearbox). This leads to unpredictable behavior including intrusion into right lane. Car on right lane collides with visible AFA. ;S3;Collision with high differential velocity.;E4;Scenario (sloped road, traffic on right lane) usually met at each deployment.;C1;AFA drifts slowly (\eg\ \SI[mode=text]{0.4}{\metre/\second} lateral) into right driving lane. Following traffic is easily able to recognize this and react appropriately (braking, avoiding).;B;SG01;Unintended and not permitted operating mode change must be prevented.
11;12;Safe Halt;Operating mode change;Unintended or not permitted transition to \coupledmode;No Hazard; ;\nohazard{Supervision works as defined in \coupledmode. Immediate operating mode change to \safehalt\ since conditions for \coupledmode\ are not met (distance to leading vehicle, transmission of odometry data of leading vehicle etc.)};S0;---;E0;---;C0;---;QM;---; 
12;13;Safe Halt;Operating mode change;Unintended or not permitted transition to \followmode;No Hazard; ;\nohazard{Due to necessary boundary conditions, \followmode\ cannot be retained. Operating mode changes back to \safehalt. };S0;---;E0;---;C0;---;QM;---; 
13;14;Safe Halt;Operating mode change;Intended and permitted transition to \manualmode\ is not executed;No Hazard; ;\nohazard{AFA still in standstill.};S0;---;E0;---;C0;---;QM;---; 
14;15;Safe Halt;Operating mode change;Intended and permitted transition to \coupledmode\ is not executed;No Hazard; ;\nohazard{AFA still in standstill.};S0;---;E0;---;C0;---;QM;---; 
15;16;Safe Halt;Operating mode change;Intended and permitted transition to \followmode\ is not executed;No Hazard; ;\nohazard{AFA still in standstill.};S0;---;E0;---;C0;---;QM;---; 
16;17;Safe Halt;Longitudinal guidance;Unintended (slow) acceleration;Hazard;Collision with traffic participants;Truck convoy on right lane.  AFA on hard shoulder starts to roll (slope, automatic gearbox). This leads to unpredictable behavior including intrusion into right lane. Truck avoids AFA, following truck touches AFA as the AFA is masked by first truck.;S2;Collision with high differential velocity, vehicles slightly touch.;E4;Scenario usually met at each deployment.;C2;AFA drifts slowly (\eg\ \SI[mode=text]{0.4}{\metre/\second} lateral) into right driving lane. Despite of masking by truck in front, following traffic is normally able to recognize this and react appropriately (braking, avoiding).;B;SG08;Unintended slow acceleration must be prevented.
16a;18;Safe Halt;Longitudinal guidance;Unintended (slow) acceleration;Hazard;Collision with traffic participants;Moving car traffic on right lane. AFA on hard shoulder starts to roll (slope, automatic gearbox). This leads to unpredictable behavior including intrusion into right lane. Car on right lane collides with visible AFA. ;S3;Collision with high differential velocity.;E4;Scenario usually met at each deployment.;C1;AFA drifts slowly (\eg\ \SI[mode=text]{0.4}{\metre/\second} lateral) into right driving lane. Following traffic is easily able to recognize this and react appropriately (braking, avoiding).;B;SG08;Unintended slow acceleration must be prevented.
17;19;Safe Halt;Longitudinal guidance;Unintended maximum deceleration;No Hazard; ;\nohazard{Maximum deceleration unproblematic, as AFA decelerates from very low velocity. Moreover, transition from \manualmode\ to \safehalt\ only possible in standstill.};S0;---;E0;---;C0;---;QM;---; 
18;20;Safe Halt;Longitudinal guidance;No stop;Hazard;Collision with traffic participants;Truck convoy on right lane. As there is no environment perception active in \safehalt, intrusion into right lane is possible. Truck avoids AFA, following truck touches AFA as the AFA is masked by first truck.;S2;Collision with high differential velocity, vehicles slightly touch.;E4;\safehalt\ active at each deployment.;C2;AFA drifts slowly (\eg\ \SI[mode=text]{0.4}{\metre/\second} lateral) into right driving lane. Despite of masking by truck in front, following traffic is normally able to recognize this and react appropriately (braking, avoiding).;B;SG09;Deceleration to standstill must be ensured.
18a;21;Safe Halt;Longitudinal guidance;No stop;Hazard;Collision with traffic participants;Moving car traffic on right lane. As there is no environment perception active in \safehalt, intrusion into right lane is possible. Car on right lane collides with visible AFA. ;S3;Collision with high differential velocity.;E4;\safehalt\ active at each deployment.;C1;AFA drifts slowly (\eg\ \SI[mode=text]{0.4}{\metre/\second} lateral) into right driving lane. Following traffic is easily able to recognize this and react appropriately (braking, avoiding).;B;SG09;Deceleration to standstill must be ensured.
19;22;Safe Halt;Longitudinal guidance;No stop;Hazard;Collision with traffic participants;Obstacle, \eg\ an emergency stopping vehicle, on hard shoulder. A person stands between vehicle and AFA. AFA collides with vehicle.;S2;Very low velocity of AFA. Person between AFA and vehicle. This is expected to lead to severe yet not fatal injuries ;E2;Obstacles on hard shoulder occur approximately once per week. A vehicle stopping between AFA and leading vehicle is even more unlikely.;C1;People between AFA and obstacle can easily react to a non-stopping AFA by stepping aside due to its low velocity. ;QM;SG09;Deceleration to standstill must be ensured.
19a;23;Safe Halt;Longitudinal guidance;No stop;Hazard;Collision with traffic participants;Obstacle, \eg\ an emergency stopping vehicle, on hard shoulder, passengers in vehicle. AFA collides with vehicle.;S0;Very low velocity of AFA. No injuries expected as people in vehicle are protected by passenger cabin.;E2;Obstacles on hard shoulder occur approximately once per week. A vehicle stopping between AFA and leading vehicle is even more unlikely.;C3;People in a stopping vehicle only have  a very small chance to react to the AFA colliding unexpectedly. Driver might press the brake pedal intuitively. ;QM;SG09;Deceleration to standstill must be ensured.
#20;24;Safe Halt;Obstacle detection;Vehicle does not react to obstacle;Hazard;Collision with traffic participants;Während des Bremsen in den Stillstand beim Wechsel aus einem der anderen beiden automatisierten Betriebsmodi ist die Obstacle detection aktiv. Bei Nicht-Erkennen eines Hindernis kann es zu einer Kollision kommen.;S1;Kollision mit niedriger Geschwindigkeit;E1;Combination of transition to \safehalt\ and simultaneous appearing obstacle very unlikely. ;C3;Vehicle in standstill cannot avoid AFA.;QM; ; 
#21;25;Safe Halt;Obstacle detection;Vehicle reacts to not existing obstacle;No Hazard; ; ;S0;Higher deceleration of AFA, yet this does not lead to hazardous event.;E0;Higher deceleration of AFA, yet this does not lead to hazardous event.;C0;Higher deceleration of AFA, yet this does not lead to hazardous event.;QM; ; 
#22;26;Coupled Mode;Detection of driver intervention;Driver intervention is not detected;keine Betrachtung; ;Test operation on hard shoulder, driver intervention is not detected. ;S0;Only for testing purposes. Driver can stop AFA pneumatically by foot brake. ;E0;Only for testing purposes. Driver can stop AFA pneumatically by foot brake. ;C0;Only for testing purposes. Driver can stop AFA pneumatically by foot brake. ;QM;SG06;Detection of driver intervention must be ensured.
#23;27;Coupled Mode;HMI;HMI displays wrong operating mode;Hazard;Collision with traffic participants;Moving traffic on right lane. In order to bypass obstacles \safehalt\ is supposed to be activated. HMI displays \safehalt, yet AFA still in \coupledmode. Operator leaves leading vehicle, in order to change to AFA. In the meantime, the leading vehicle starts to bypass the obstacle. AFA also starts driving. In Coupled Mode, perception of lane markings is not active. Leading vehicle accelerates v>\SI[mode=text]{10}{\km\per\hour} and changes to right lane. AFA stops as intended, in case the distance to leading vehicle exceeds its tolerance range. AFA stops partially on right lane. Vehicle on right lane collides with visible AFA. ;S3;Collision with high differential velocity.;E4;Bypass of obstacles and barriers (\eg\ bridges) on each deployment.;C0;Leading vehicle only changes to right lane if gap is large enough. Warning device is active. Thus, this scenario is very easy to handle by traffic participants.;QM;SG07;Display of actual operating mode in HMI must be ensured.
24;28;Coupled Mode;Operating mode change;Unintended or not permitted transition to \manualmode;Hazard;Collision with traffic participants;Truck convoy on right lane. AFA on acceleration or deceleration lane. \manualmode\ without human driver. This leads to unpredictable behavior including intrusion into right lane. Truck avoids AFA, following truck touches AFA as the AFA is masked by first truck.;S2;Collision with high differential velocity, vehicles slightly touch.;E4;Passing acceleration and deceleration lanes occurs on each deployment. ;C2;AFA drifts slowly (\eg\ \SI[mode=text]{0.4}{\metre/\second} lateral) into right driving lane. Despite of masking by truck in front, following traffic is normally able to recognize this and react appropriately (braking, avoiding).;B;SG01;Unintended and not permitted operating mode change must be prevented.
24a;29;Coupled Mode;Operating mode change;Unintended or not permitted transition to \manualmode;Hazard;Collision with traffic participants;Moving car traffic on right lane. AFA on acceleration or deceleration lane. \manualmode\ without human driver. This leads to unpredictable behavior including intrusion into right lane.  Car on right lane collides with visible AFA. ;S3;Collision with high differential velocity.;E4;Passing acceleration and deceleration lanes occurs on each deployment. ;C1;AFA drifts slowly (\eg\ \SI[mode=text]{0.4}{\metre/\second} lateral) into right driving lane. Following traffic is easily able to recognize this and react appropriately (braking, avoiding).;B;SG01;Unintended and not permitted operating mode change must be prevented.
25;30;Coupled Mode;Operating mode change;Unintended or not permitted transition to \safehalt;Hazard;Collision with traffic participants;Moving traffic on right lane. AFA stops on acceleration or deceleration lane. Vehicles entering or leaving the motorway collide with AFA.;S2;Rear-end collision with reduced velocity;E4;Passing acceleration and deceleration lanes occurs on each deployment. ;C0;Driver of vehicle changing to deceleration lane is already braking or ready for braking. Vehicle on acceleration lane in general has moderate velocity.;QM;SG01;Unintended and not permitted operating mode change must be prevented.
26;31;Coupled Mode;Operating mode change;Unintended or not permitted transition to \followmode;Hazard;Collision with traffic participants;Moving traffic on right lane. AFA stops on acceleration or deceleration lane in order to build up required distance for \followmode. Vehicles entering or leaving the motorway collide with AFA.;S2;Rear-end collision with reduced velocity;E4;Passing acceleration and deceleration lanes occurs on each deployment. ;C0;Driver of vehicle changing to deceleration lane is already braking or ready for braking. Vehicle on acceleration lane in general has moderate velocity.;QM;SG01;Unintended and not permitted operating mode change must be prevented.
27;32;Coupled Mode;Operating mode change;Intended and permitted transition to \manualmode\ is not executed;keine Betrachtung; ;\nohazard{Only for testing purposes. Driver can stop AFA pneumatically by foot brake.};S0;---;E0;---;C0;---;QM;---; 
28;33;Coupled Mode;Operating mode change;Intended and permitted transition to \safehalt\ is not executed;Hazard;Collision with traffic participants;Truck convoy on right lane. Operating mode change to \safehalt\ when exceeding functional system boundaries is not executed. This leads to unpredictable behavior including intrusion into right lane. Truck avoids AFA, following truck touches AFA as the AFA is masked by first truck.;S2;Collision with high differential velocity, vehicles slightly touch.;E4;Passing acceleration and deceleration lanes occurs on each deployment. ;C2;AFA drifts slowly (\eg\ \SI[mode=text]{0.4}{\metre/\second} lateral) into right driving lane. Despite of masking by truck in front, following traffic is normally able to recognize this and react appropriately (braking, avoiding).;B;SG02;Intended and permitted operating mode change must be ensured.
28a;34;Coupled Mode;Operating mode change;Intended and permitted transition to \safehalt\ is not executed;Hazard;Collision with traffic participants;Moving car traffic on right lane. Operating mode change to \safehalt\ when exceeding functional system boundaries is not executed. This leads to unpredictable behavior including intrusion into right lane. Car on right lane collides with visible AFA. ;S3;Collision with high differential velocity.;E4;Passing acceleration and deceleration lanes occurs on each deployment. ;C1;AFA drifts slowly (\eg\ \SI[mode=text]{0.4}{\metre/\second} lateral) into right driving lane. Following traffic is easily able to recognize this and react appropriately (braking, avoiding).;B;SG02;Intended and permitted operating mode change must be ensured.
#29;35;Coupled Mode;Longitudinal and lateral guidance;Vehicle does not follow leading vehicle;Hazard;Collision with traffic participants;Fließender Verkehr auf dem benachbarten Fahrstreifen. Kritisch ist hier die Lateral guidance. AFA schwenkt in den fließenden Verkehr und kollidiert mit einem Verkehrsteilnehmer.;S3;Collision with high differential velocity.;E4;Passing acceleration and deceleration lanes occurs on each deployment. ;C2;Plötzlich auf den rechten Fahrstreifen wechselndes AFA bei geringem Abstand zum Fahrstreifen nur schwer für nachfolgenden Verkehr zu beherrschen. ;C; ; 
30;36;Coupled Mode;Longitudinal and lateral guidance;Vehicle does not follow in defined distance (tolerance range lateral or longitudinal);Hazard;Collision with traffic participants;Truck convoy on right lane. AFA follows leading vehicle with lateral and longitudinal offsets which exceed the tolerance ranges. AFA is partially driving on right lane. Truck avoids AFA, following truck touches AFA as the AFA is masked by first truck.;S2;Collision with high differential velocity, vehicles slightly touch.;E4;Passing acceleration and deceleration lanes occurs on each deployment. ;C0;Traffic participants can control the AFA protruding into the right lane as the warning device is active as well as lateral and Longitudinal guidance function as intended apart from the lateral offset.;QM;SG10;Leaving tolerance ranges must trigger operating mode change to \safehalt.
30a;37;Coupled Mode;Longitudinal and lateral guidance;Vehicle does not follow in defined distance (tolerance range lateral or longitudinal);Hazard;Collision with traffic participants;Moving car traffic on right lane. AFA follows leading vehicle with lateral and longitudinal offsets which exceed the tolerance ranges. AFA is partially driving on right lane. Car on right lane collides with visible AFA. ;S3;Collision with high differential velocity.;E4;Passing acceleration and deceleration lanes occurs on each deployment. ;C0;Traffic participants can control the AFA protruding into the right lane as the warning device is active as well as lateral and Longitudinal guidance function as intended apart from the lateral offset.;QM;SG10;Leaving tolerance ranges must trigger operating mode change to \safehalt.
31;38;Coupled Mode;Longitudinal guidance;Vehicle exceeds maximum speed of \SI[mode=text]{12}{\km\per\hour};Hazard;Collision with traffic participants;Truck convoy on right lane. The functional components are designed for velocities up to \SI[mode=text]{12}{\km\per\hour}. AFA drives at not excessively higher velocity. AFA exceeds functional system boundaries. This leads to unpredictable behavior including intrusion into right lane (\eg\ oscillating steering angle control). Truck avoids AFA, following truck touches AFA as the AFA is masked by first truck.;S2;Collision with high differential velocity, vehicles slightly touch.;E4;Passing acceleration and deceleration lanes occurs on each deployment. ;C2;AFA drifts slowly (\eg\ \SI[mode=text]{0.4}{\metre/\second} lateral) into right driving lane. Despite of masking by truck in front, following traffic is normally able to recognize this and react appropriately (braking, avoiding).;B;SG11;Maximum velocity must not be exceeded.
31a;39;Coupled Mode;Longitudinal guidance;Vehicle exceeds maximum speed of \SI[mode=text]{12}{\km\per\hour};Hazard;Collision with traffic participants;Moving car traffic on right lane. The functional components are designed for velocities up to \SI[mode=text]{12}{\km\per\hour}. AFA drives at not excessively higher velocity. AFA exceeds functional system boundaries. This leads to unpredictable behavior including intrusion into right lane (\eg\ oscillating steering angle control). Car on right lane collides with visible AFA. ;S3;Collision with high differential velocity.;E4;Passing acceleration and deceleration lanes occurs on each deployment. ;C1;AFA drifts slowly (\eg\ \SI[mode=text]{0.4}{\metre/\second} lateral) into right driving lane. Following traffic is easily able to recognize this and react appropriately (braking, avoiding).;B;SG11;Maximum velocity must not be exceeded.
#32;40;Coupled Mode;Obstacle detection;Vehicle does not react to obstacle;Hazard;Collision with traffic participants;Anderer Verkehrsteilnehmer schert zwischen AFA und Führungsfahrzeug ein. AFA reagiert nicht auf weiter verzögernden Verkehrsteilnehmer.;S1;Rear-end collision with reduced velocity;E0;Einscheren anderer Verkehrsteilnehmer unwahrscheinlich auf Grund des geringen Abstands und der hohen Differenzgeschwindigkeiten.;C1;People between AFA and obstacle can easily react to a non-stopping AFA by stepping aside due to its low velocity. ;QM;---; 
33;41;Coupled Mode;Longitudinal guidance;Unintended deceleration;Hazard;Collision with traffic participants;Moving traffic on right lane. AFA stops on acceleration or deceleration lane. Vehicles entering or leaving the motorway collide with AFA.;S2;Rear-end collision with reduced velocity;E4;Passing acceleration and deceleration lanes occurs on each deployment. ;C0;Driver of vehicle changing to deceleration lane is already braking or ready for braking. Vehicle on acceleration lane in general has moderate velocity.;QM;SG04;Unintended anti-lock brake actuation must be prevented.
34;42;Coupled Mode;Radio communication;Vehicle drives without radio communication;Hazard;Collision with traffic participants;AFA stops due to inconsistent data of radio communication and environment perception on acceleration or deceleration lane. Vehicles entering or leaving the motorway collide with AFA.;S2;Rear-end collision with reduced velocity;E4;Passing acceleration and deceleration lanes occurs on each deployment. ;C0;Driver of vehicle changing to deceleration lane is already braking or ready for braking. Vehicle on acceleration lane in general has moderate velocity.;QM;SG10;Leaving tolerance ranges must trigger operating mode change to \safehalt.
35;43;Coupled Mode;Lateral guidance;Steering angle change beyond maximum specification (angle \& change rate);Hazard;Collision with traffic participants;AFA drifts with up to maximum possible yaw rate into right lane. ;S3;Collision with high differential velocity.;E4;Passing acceleration and deceleration lanes occurs on each deployment. ;C3;AFA drifts quickly (\eg\ $\gg$\SI[mode=text]{0.4}{\metre/\second} lateral) into the right lane. It follows a circular arc to the guardrail on the left of the left lane. This is difficult to control by traffic participants. ;D;SG03;Unintended maximum steering actuation must be prevented.
36;44;Follow Mode;Detection of driver intervention;Driver intervention is not detected;keine Betrachtung; ;Test operation on hard shoulder, driver intervention is not detected. ;S0;Only for testing purposes. Driver can stop AFA pneumatically by foot brake. ;E0;Only for testing purposes. Driver can stop AFA pneumatically by foot brake. ;C0;Only for testing purposes. Driver can stop AFA pneumatically by foot brake. ;QM;SG06;Detection of driver intervention must be ensured.
37;45;Follow Mode;HMI;HMI displays wrong operating mode;Hazard;Collision with traffic participants;Truck convoy on right lane. Leading vehicle in standstill, AFA in \followmode. HMI displays \safehalt\ or \coupledmode. AFA starts delayed to follow leading vehicle and enters acceleration or deceleration lane. AFA exceeds functional system boundaries. This leads to unpredictable behavior including intrusion into right lane. Truck avoids AFA, following truck touches AFA as the AFA is masked by first truck.;S2;Collision with high differential velocity, vehicles slightly touch.;E4;Operation on hard shoulder occurs on each deployment.;C1;AFA drifts slowly (\eg\ \SI[mode=text]{0.4}{\metre/\second} lateral) into right driving lane. Despite of masking by truck in front, following traffic is normally able to recognize this and react appropriately (braking, avoiding).;A;SG07;Display of actual operating mode in HMI must be ensured.
37a;46;Follow Mode;HMI;HMI displays wrong operating mode;Hazard;Collision with traffic participants;Moving car traffic on right lane. Leading vehicle in standstill, AFA in \followmode. HMI displays \safehalt\ or \coupledmode. AFA starts delayed to follow leading vehicle and enters acceleration or deceleration lane. AFA exceeds functional system boundaries. This leads to unpredictable behavior including intrusion into right lane. Car on right lane collides with visible AFA. ;S2;Collision with high differential velocity.;E4;Operation on hard shoulder occurs on each deployment.;C1;AFA drifts slowly (\eg\ \SI[mode=text]{0.4}{\metre/\second} lateral) into right driving lane. Following traffic is easily able to recognize this and react appropriately (braking, avoiding).;A;SG07;Display of actual operating mode in HMI must be ensured.
38;47;Follow Mode;Operating mode change;Unintended or not permitted transition to \manualmode;Hazard;Collision with traffic participants;Truck convoy on right lane.  \manualmode\ without driver. This leads to unpredictable behavior including intrusion into right lane. Truck avoids AFA, following truck touches AFA as the AFA is masked by first truck.;S2;Collision with high differential velocity, vehicles slightly touch.;E4;Operation on hard shoulder occurs on each deployment.;C2;AFA drifts slowly (\eg\ \SI[mode=text]{0.4}{\metre/\second} lateral) into right driving lane. Despite of masking by truck in front, following traffic is normally able to recognize this and react appropriately (braking, avoiding).;B;SG01;Unintended and not permitted operating mode change must be prevented.
38a;48;Follow Mode;Operating mode change;Unintended or not permitted transition to \manualmode;Hazard;Collision with traffic participants;Moving car traffic on right lane. \manualmode\ without driver. This leads to unpredictable behavior including intrusion into right lane. Car on right lane collides with visible AFA. ;S3;Collision with high differential velocity.;E4;Operation on hard shoulder occurs on each deployment.;C1;AFA drifts slowly (\eg\ \SI[mode=text]{0.4}{\metre/\second} lateral) into right driving lane. Following traffic is easily able to recognize this and react appropriately (braking, avoiding).;B;SG01;Unintended and not permitted operating mode change must be prevented.
#39;49;Follow Mode;Operating mode change;Unintended or not permitted transition to \safehalt;No Hazard; ; ;S0;AFA stops on hard shoulder.;E0;AFA stops on hard shoulder.;C0;AFA stops on hard shoulder.;QM;---; 
#40;50;Follow Mode;Operating mode change;Unintended or not permitted transition to \coupledmode;No Hazard; ; ;S0;Distance to leading vehicle to large. Immediate transition to \safehalt.;E0;Distance to leading vehicle to large. Immediate transition to \safehalt.;C0;Distance to leading vehicle to large. Immediate transition to \safehalt.;QM;---; 
41;51;Follow Mode;Follow hard shoulder;Vehicle does not follow hard shoulder;Hazard;Collision with traffic participants;Truck convoy on right lane. AFA intrudes right lane. Truck avoids AFA, following truck touches AFA as the AFA is masked by first truck.;S2;Collision with high differential velocity, vehicles slightly touch.;E4;Operation on hard shoulder occurs on each deployment.;C2;AFA drifts slowly (\eg\ \SI[mode=text]{0.4}{\metre/\second} lateral) into right driving lane. Despite of masking by truck in front, following traffic is normally able to recognize this and react appropriately (braking, avoiding).;B;SG12;Overrunning hard shoulder markings must be prevented.
41a;52;Follow Mode;Follow hard shoulder;Vehicle does not follow hard shoulder;Hazard;Collision with traffic participants;Moving car traffic on right lane. AFA intrudes right lane. Car on right lane collides with visible AFA. ;S3;Collision with high differential velocity.;E4;Operation on hard shoulder occurs on each deployment.;C1;AFA drifts slowly (\eg\ \SI[mode=text]{0.4}{\metre/\second} lateral) into right driving lane. Following traffic is easily able to recognize this and react appropriately (braking, avoiding).;B;SG12;Overrunning hard shoulder markings must be prevented.
42;53;Follow Mode;Keep defined distance;Vehicle does not follow in defined tolerance range;No Hazard; ;\nohazard{AFA continues following hard shoulder based on lane marking. Obstacle detection functions. AFA stops if distance to leading vehicle is too large (leading vehicle out of sight, interruption of radio communication).};S0;---;E0;---;C0;---;QM;---; 
43;54;Follow Mode;Obstacle detection;Vehicle does not react to obstacle;Hazard;Collision with traffic participants;Obstacle, \eg\ an emergency stopping vehicle, on hard shoulder. A person stands between vehicle and AFA. AFA collides with vehicle.;S2;Very low velocity of AFA. Person between AFA and vehicle. This is expected to lead to severe yet not fatal injuries ;E2;Obstacles on hard shoulder occur approximately once per week. A vehicle stopping between AFA and leading vehicle is even more unlikely.;C1;People between AFA and obstacle can easily react to a non-stopping AFA by stepping aside due to its low velocity. ;QM;SG13;Detection of and reaction to (deceleration to standstill) relevant obstacles must be ensured. 
43a;55;Follow Mode;Obstacle detection;Vehicle does not react to obstacle;Hazard;Collision with traffic participants;Obstacle, \eg\ an emergency stopping vehicle, on hard shoulder, passengers in vehicle. AFA collides with vehicle.;S0;Very low velocity of AFA. No injuries expected as people in vehicle are protected by passenger cabin.;E2;Obstacles on hard shoulder occur approximately once per week. A vehicle stopping between AFA and leading vehicle is even more unlikely.;C3;People in a stopping vehicle only have  a very small chance to react to the AFA colliding unexpectedly. Driver might press the brake pedal intuitively. ;QM;SG13;Detection of and reaction to (deceleration to standstill) relevant obstacles must be ensured. 
44;56;Follow Mode;Longitudinal guidance;Unintended deceleration;No Hazard; ;\nohazard{AFA stops on hard shoulder with active warning device and transitions to \safehalt. };S0;---;E0;---;C0;---;QM;---; 
45;57;Follow Mode;Perceive leading vehicle;Vehicle keeps distance to wrong object;Hazard;Collision with traffic participants;Truck convoy on right lane. AFA follows wrong leading vehicle which does not stop in front of acceleration or deceleration lanes. AFA exceeds functional system boundaries. This leads to unpredictable behavior including intrusion into right lane. Truck avoids AFA, following truck touches AFA as the AFA is masked by first truck.;S2;Collision with high differential velocity, vehicles slightly touch.;E1;Vehicle driving on hard shoulder for a longer period of time and with velocity $\leq$ \SI[mode=text]{10}{\km\per\hour} occurs very rarely. ;C2;AFA drifts slowly (\eg\ \SI[mode=text]{0.4}{\metre/\second} lateral) into right driving lane. Despite of masking by truck in front, following traffic is normally able to recognize this and react appropriately (braking, avoiding).;QM;SG14;Identification of leading vehicle must be ensured.
45a;58;Follow Mode;Perceive leading vehicle;Vehicle keeps distance to wrong object;Hazard;Collision with traffic participants;Moving car traffic on right lane. AFA follows wrong leading vehicle which does not stop in front of acceleration or deceleration lanes. AFA exceeds functional system boundaries. This leads to unpredictable behavior including intrusion into right lane. Car on right lane collides with visible AFA. ;S3;Collision with high differential velocity.;E1;Vehicle driving on hard shoulder for a longer period of time and with velocity $\leq$ \SI[mode=text]{10}{\km\per\hour} occurs very rarely. ;C1;AFA drifts slowly (\eg\ \SI[mode=text]{0.4}{\metre/\second} lateral) into right driving lane. Following traffic is easily able to recognize this and react appropriately (braking, avoiding).;QM;SG14;Identification of leading vehicle must be ensured.
46;59;Follow Mode;Perceive leading vehicle;Vehicle follows hard shoulder without leading vehicle;Hazard;Collision with traffic participants;Truck convoy on right lane as well as vehicles driving on acceleration and deceleration lane. AFA does not detect begin of acceleration or deceleration lane. Thus, it exceeds its functional system boundaries. This leads to unpredictable behavior including intrusion into right lane.;S2;Collision with high differential velocity, vehicles slightly touch.;E0;Operating instructions prohibit activation of automated operation without leading vehicle.;C2;AFA drifts slowly (\eg\ \SI[mode=text]{0.4}{\metre/\second} lateral) into right driving lane. Despite of masking by truck in front, following traffic is normally able to recognize this and react appropriately (braking, avoiding).;QM;SG15;Detection of missing leading vehicle and operating mode change to \safehalt\ must be ensured.
46a;60;Follow Mode;Perceive leading vehicle;Vehicle follows hard shoulder without leading vehicle;Hazard;Collision with traffic participants;Moving car traffic on right lane as well as vehicles driving on acceleration and deceleration lane. AFA does not detect begin of acceleration or deceleration lane. Thus, it exceeds its functional system boundaries. This leads to unpredictable behavior including intrusion into right lane.;S3;Collision with high differential velocity.;E0;Operating instructions prohibit activation of automated operation without leading vehicle.;C1;AFA drifts slowly (\eg\ \SI[mode=text]{0.4}{\metre/\second} lateral) into right driving lane. Following traffic is easily able to recognize this and react appropriately (braking, avoiding).;QM;SG15;Detection of missing leading vehicle and operating mode change to \safehalt\ must be ensured.
47;61;Follow Mode;Radio communication;Vehicle leaves range of radio communication;No Hazard; ;\nohazard{Interruption of radio communication causes transition to \safehalt. AFA stops on hard shoulder with active warning device.};S0;---;E0;---;C0;---;QM;---; 
48;62;Follow Mode;Operating mode change;Intended and permitted transition to \safehalt\ is not executed;Hazard;Collision with traffic participants;Truck convoy on right lane. AFA must transit to \safehalt, \eg\ due to exceeding a functional system boundary. transition to \safehalt\ does not function. AFA exceeds functional system boundaries. This leads to unpredictable behavior including intrusion into right lane (\eg\ oscillating steering angle control). Truck avoids AFA, following truck touches AFA as the AFA is masked by first truck.;S2;Collision with high differential velocity, vehicles slightly touch.;E4;Operation on hard shoulder occurs on each deployment.;C2;AFA drifts slowly (\eg\ \SI[mode=text]{0.4}{\metre/\second} lateral) into right driving lane. Despite of masking by truck in front, following traffic is normally able to recognize this and react appropriately (braking, avoiding).;B;SG02;Intended and permitted operating mode change must be ensured.
48a;63;Follow Mode;Operating mode change;Intended and permitted transition to \safehalt\ is not executed;Hazard;Collision with traffic participants;Moving car traffic on right lane. AFA must transit to \safehalt, \eg\ due to exceeding a functional system boundary. transition to \safehalt\ does not function. AFA exceeds functional system boundaries. This leads to unpredictable behavior including intrusion into right lane (\eg\ oscillating steering angle control). Car on right lane collides with visible AFA. ;S3;Collision with high differential velocity.;E4;Operation on hard shoulder occurs on each deployment.;C1;AFA drifts slowly (\eg\ \SI[mode=text]{0.4}{\metre/\second} lateral) into right driving lane. Following traffic is easily able to recognize this and react appropriately (braking, avoiding).;B;SG02;Intended and permitted operating mode change must be ensured.
49;64;Follow Mode;Operating mode change;Intended and permitted transition to \manualmode\ is not executed;keine Betrachtung; ;\nohazard{Only for testing purposes. Driver can stop AFA pneumatically by foot brake.};S0;---;E0;---;C0;---;QM;---; 
50;65;Follow Mode;Radio communication;Vehicle drives without radio communication;No Hazard; ;\nohazard{Detection of lane markings and obstacles as well as HMI function as intended.  AFA stops when leading vehicle stops according work instructions before passing acceleration or deceleration lanes. Then, malfunction becomes obvious by missing transition to \coupledmode.};S0;---;E0;---;C0;---;QM;---; 
51;66;Follow Mode;Longitudinal guidance;Vehicle exceeds maximum speed of \SI[mode=text]{12}{\km\per\hour};Hazard;Collision with traffic participants;Truck convoy on right lane. The functional components are designed for velocities up to \SI[mode=text]{12}{\km\per\hour}. AFA drives at not excessively higher velocity. AFA exceeds functional system boundaries. This leads to unpredictable behavior including intrusion into right lane (\eg\ oscillating steering angle control). Truck avoids AFA, following truck touches AFA as the AFA is masked by first truck.;S2;Collision with high differential velocity, vehicles slightly touch.;E4;Operation on hard shoulder occurs on each deployment.;C2;AFA drifts slowly (\eg\ \SI[mode=text]{0.4}{\metre/\second} lateral) into right driving lane. Despite of masking by truck in front, following traffic is normally able to recognize this and react appropriately (braking, avoiding).;B;SG11;Maximum velocity must not be exceeded.
51a;67;Follow Mode;Longitudinal guidance;Vehicle exceeds maximum speed of \SI[mode=text]{12}{\km\per\hour};Hazard;Collision with traffic participants;Moving car traffic on right lane. The functional components are designed for velocities up to \SI[mode=text]{12}{\km\per\hour}. AFA drives at not excessively higher velocity. AFA exceeds functional system boundaries. This leads to unpredictable behavior including intrusion into right lane (\eg\ oscillating steering angle control). Car on right lane collides with visible AFA. ;S3;Collision with high differential velocity.;E4;Operation on hard shoulder occurs on each deployment.;C1;AFA drifts slowly (\eg\ \SI[mode=text]{0.4}{\metre/\second} lateral) into right driving lane. Following traffic is easily able to recognize this and react appropriately (braking, avoiding).;B;SG11;Maximum velocity must not be exceeded.
52;68;Follow Mode;Lateral guidance;Steering angle change beyond maximum specification (angle \& change rate);Hazard;Collision with traffic participants;AFA drifts with up to maximum possible yaw rate into right lane. ;S3;Collision with high differential velocity.;E4;Operation on hard shoulder occurs on each deployment.;C3;AFA drifts quickly (\eg\ $\gg$\SI[mode=text]{0.4}{\metre/\second} lateral) into the right lane. It follows a circular arc to the guardrail on the left of the left lane. This is difficult to control by traffic participants. ;D;SG03;Unintended maximum steering actuation must be prevented.
\end{filecontents*}
\pgfplotstabletypeset[
	empty header,
	begin table=\begin{longtable},
	columns={
		 Original ID,
		 Operating mode, 
		 Function,
		 Malfunction,
		 Hazardous Event,
		 Severity,
		 Severity Rationale,
		 Exposure,
		 Exposure Rationale,
		 Controllability,
		 Controllability Rationale,
		 ASIL Classification,
		 Safety Goal ID
	},
	display columns/0/.style={
		column type= >{\centering\arraybackslash}m{\IDwidth}},
	display columns/1/.style={
		column type= >{\centering\arraybackslash}m{\modeWidth}},
	display columns/2/.style={
		column type= >{\centering\arraybackslash}m{\functionWidth}},
	display columns/3/.style={
		column type= >{\centering\arraybackslash}m{\malfunctionWidth}},
	display columns/4/.style={
		column type= >{\centering\arraybackslash}m{\hazardousEventWidth}},
	display columns/5/.style={
		column type= >{\centering\arraybackslash}m{\SECWidth}},
	display columns/6/.style={
		column type= >{\centering\arraybackslash}m{\SRationaleWidth}},
	display columns/7/.style={
		column type= >{\centering\arraybackslash}m{\SECWidth}},
	display columns/8/.style={
		column type= >{\centering\arraybackslash}m{\ERationaleWidth}},
	display columns/9/.style={
		column type= >{\centering\arraybackslash}m{\SECWidth}},
	display columns/10/.style={
		column type= >{\centering\arraybackslash}m{\CRationaleWidth}},
	display columns/11/.style={
		column type= >{\centering\arraybackslash}m{\ASILWidth}},
	display columns/12/.style={
		column type= >{\centering\arraybackslash}m{\SGWidth}},
	every first row/.append style={before row={%
			\toprule
			 ID 
			&Operating Mode
			&Function 
			&Malfunction 
			&Hazardous Scenario and Consequence 
			&S 
			&Rationale 
			&E 
			&Rationale 
			&C	
			&Rationale 
			&A 
			&SG
			\\ \midrule    
			\endfirsthead
			\multicolumn{13}{c}%
			{{\footnotesize TABLE \thetable : Continued from previous page}} \\
			\toprule 
			 ID 
			&Operating Mode
			&Function 
			&Malfunction 
			&Hazardous Scenario and Consequence 
			&S 
			&Rationale 
			&E 
			&Rationale 
			&C	
			&Rationale 
			&A 
			&SG
			\\  \midrule  
			\endhead
			\midrule 
			\multicolumn{3}{l}{{Continued on next page}}&
			\bottomtext \\ 
			\bottomrule
			\endfoot
			\midrule
			\multicolumn{3}{l}{{Table concluded}}&
			\bottomtext	\\ 
			\bottomrule
			\endlastfoot
		}
	},
	postproc cell content/.code={
		\ifodd\pgfplotstablerow\relax
		\else
			\ifnum\pgfplotstablecol>-1
				\pgfkeysalso{@cell content={\cellcolor{lightgrey}#1}}%
			\fi
		\fi
	},	
	end table=\end{longtable},
	col sep=semicolon,
	string type
]
{tables/hara.csv}

\end{landscape}


\end{document}